\documentclass{sigchi}
\usepackage{authblk}
\usepackage[T1]{fontenc}   
\usepackage{txfonts}
\usepackage{mathptmx}
\usepackage{hyperref}
\usepackage{booktabs}
\usepackage{textcomp}
\usepackage{microtype}        
\usepackage{ccicons}          

\usepackage{helvet}
\usepackage{courier}
\usepackage{bm}
\usepackage{balance}  
\usepackage{graphicx}
\usepackage{url}      
\usepackage{multirow}
\usepackage{color}
\usepackage{xcolor}
\usepackage{soul}
\usepackage{longtable}
\usepackage{multicol}
\usepackage{lipsum}
\usepackage{paralist}
\usepackage[tight,footnotesize]{subfigure}
\usepackage{etoolbox}

\def\plaintitle{SIGCHI Conference Proceedings Format}

\def\emptyauthor{}
\def\plainkeywords{Authors' choice; of terms; separated; by
  semicolons; include commas, within terms only; required.}

\newcommand{\para}[1]{{\vspace{2pt} \noindent \em #1 \hspace{1pt}}}

\newenvironment{packed_itemize}{
\begin{list}{\labelitemi}{\leftmargin=0.5em}
  \setlength{\itemsep}{2pt}
  \setlength{\parskip}{0pt}
  \setlength{\parsep}{0pt}
  \setlength{\headsep}{0pt}
  \setlength{\topskip}{0pt}
  \setlength{\topmargin}{0pt}
  \setlength{\topsep}{0pt}
  \setlength{\partopsep}{0pt}
}{\end{list}}

\makeatletter
\def\url@leostyle{%
  \@ifundefined{selectfont}{
    \def\UrlFont{\sf}
  }{
    \def\UrlFont{\small\bf\ttfamily}
  }}
\makeatother
\def\pprw{8.5in}
\def\pprh{11in}

\definecolor{linkColor}{RGB}{6,125,233}
\hypersetup{%
  pdftitle={\plaintitle},
  pdfauthor={\emptyauthor},
  pdfkeywords={\plainkeywords},
  pdfdisplaydoctitle=true, 
  bookmarksnumbered,
  pdfstartview={FitH},
  colorlinks,
  citecolor=black,
  filecolor=black,
  linkcolor=black,
  urlcolor=linkColor,
  breaklinks=true,
  hypertexnames=false
}

\makeatletter
\def\@copyrightspace{\relax}
\makeatother

\begin{document}

\title{Towards Monetary Incentives in Social Q\&A Services}
 \author[*]{Steve T.K. Jan}
 \author[*]{Chun	Wang}
 \author[$\dag$]{Qing Zhang}
 \author[*]{Gang Wang}
 \affil[*]{Department of Computer Science, College of Engineering, Virginia Tech}
 \affil[$\dag$]{Instructional Design and Technology, School of Education, Virginia Tech}

\maketitle





\begin{abstract}

Community-based question answering (CQA) services are facing
key challenges to motivate domain experts to provide timely
answers. Recently, CQA services are
exploring new incentive models to engage experts and celebrities by
allowing them to set a price on their answers. In this paper, we
perform a data-driven analysis on two emerging {\em payment-based} CQA
systems: Fenda (China) and Whale (US). 
By analyzing a large dataset of 220K questions (worth 1 million USD
collectively), we examine how monetary incentives affect different
players in the system. We find that, while monetary incentive enables
quick answers from experts, it also drives certain users to
aggressively game the system for profits. In addition, in this
supplier-driven marketplace, users need to proactively adjust their
price to make profits. Famous people are unwilling to lower their price,
which in turn hurts their income and engagement over time. Finally, we
discuss the key implications to future CQA design.

\end{abstract}




\section{Introduction}
\label{sec:intro}
The success of community based question answering (CQA) services
depends on high-quality content from users, particularly from domain
experts. With highly engaging experts, services like Quora and
StackOverflow attract hundreds of millions of visitors
worldwide~\cite{quora16}. However, for most CQA systems, domain
experts are answering questions {\em voluntarily} for free. As the
question volume keeps growing, it becomes difficult to draw experts'
attention to a particular question, let alone getting answers
on-demand~\cite{srba2015stack}. 
 
To motivate experts, one possible direction is to introduce monetary
incentives~\cite{hsieh2009mimir}. Recently, Quora started a beta test
on ``knowledge prize'', which allows users to put cash rewards on their
questions. While
Quora is slowly accumulating interested users (less than 10 paid
answers per month), another payment-based service called {\em
  Fenda}\footnote{\url{http://fd.zaih.com/fenda}} is rising quickly in
China. Fenda is a social network app that connects users to
well-known domain experts and celebrities to ask questions with
payments. Launched in May 2016, Fenda quickly gained 10 million
registered users, 500K paid questions, and 2 million US dollar revenue
in just two months~\cite{news1-fenda16}.

Fenda leads a new wave of payment-based CQA services that socially
engage users with real-world domain experts. Similar services are
emerging in China (Zhihu, Weibo QA) and US (Whale, Campfire.fm). The
involvement of verified experts differs them from earlier
payment-based CQA services (most defunct now) that were built on an
anonymous crowd such as Mahalo Answers and
ChaCha~\cite{chen2010knowledge,hsieh2010pay,lee2013analyzing}. 

\begin{figure}
         \begin{center}
      	\begin{minipage}[t]{0.45\textwidth}
   	\includegraphics[width=0.99\textwidth]{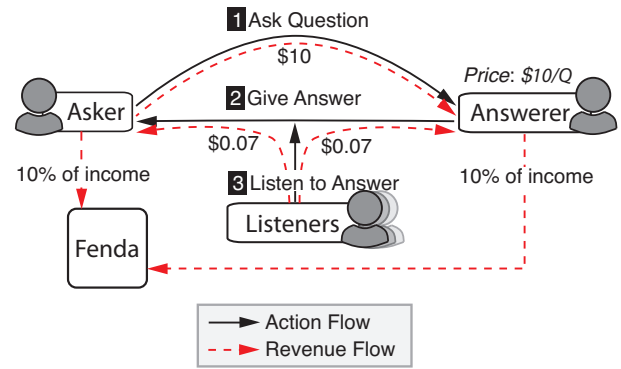}
	\end{minipage}
\caption{Fenda QA system: a user can ask another user a question by paying her (price set by the answerer). Other users need to pay a small amount to listen to the answer (\$0.14), which will be split evenly between the asker and the answerer. Fenda charges 10\% commission fee.}
	\label{fig:fenda}
  	\vspace{-0.18in}
\end{center}
\end{figure}

{\em So, is monetary incentive the solution to strong expert
  engagement in CQA systems? How does monetary incentive affect the
  behavior of different players in the system and their over-time
  engagement?} These questions are critical for payment-based CQA
design, and platforms like Fenda provide a unique opportunity to study them. First,
Fenda is the first supplier-driven CQA marketplace, where answerers
(experts) set their own price. Users ask questions to a specific
person instead of an anonymous crowd using payments. In addition,
Fenda's incentive model not only rewards answerers, but also those who
ask good questions. After a question is answered, other users need to
pay a small amount (\$0.14) to listen to the answer. This money is
split evenly between the asker and the answerer
(Figure~\ref{fig:fenda}). A good question may attract enough listeners
to compensate the initial question fee. 

In this paper, we describe our experience and findings in an effort to
understand the impact of monetary incentives on CQA systems, through a
detailed measurement of Fenda (China) and a similar system Whale\footnote{\url{https://askwhale.com/}}
(US). We collected a dataset of 88,540 users and 212,082 answers
from Fenda (two months in 2016), and 1,419 users and 9,199 answers
from Whale (6 months during 2016--2017), involving more than 1
million USD transactions. Given the drastic differences between
payment-based CQA systems and mainstream systems such as Quora and
StackOverflow, our study has significant implications for the future
direction of CQA design. 


Our study has a number of key findings. 
\begin{packed_itemize}
\item {\em First}, we seek to understand the effectiveness of monetary
  incentives to engage domain experts. Our result shows this attempt
  is successful. Both Fenda and Whale attract a small group of
  high-profile experts and celebrities who make significant
  contributions to the CQA community. For example, Fenda experts count for
  0.5\% of the user population, but have contributed a quarter of all
  answers and driven nearly half of the financial revenue.

\item {\em Second}, we analyze how the incentive model affects user
  behavior, and find a mixed effect. On the
  positive side, monetary incentive enables quick answers (average
  delay 10--23 hours) and motivates users to ask good questions (40\%
  of the Fenda questions successfully drew enough interested audience to
  cover the askers' cost). However, we did find a small
  number of manipulative users including ``bounty hunters'' who
  aggressively asked questions to make money from listeners, and
  ``collusive users'' who work together to manipulate their perceived
  popularity. 


\item {\em Third}, we study the dynamics between money and user
  engagement over time. In a supplier-driven CQA marketplace, users
  need to set the price of their answers. We find different pricing
  strategies of users have distinct impacts on their own engagement
  level. Users who proactively adjust their price are more likely to
  increase income and engagement level. Certain highly famous people,
  however, are unwilling to lower their price, which in turn hurts
  their income and social engagement.

\end{packed_itemize}

To the best of our knowledge, this is the first empirical study on
supplier-driven CQA marketplaces. Our study provides practical guidelines for other arising
payment-based CQA services (Quora knowledge
prize, Zhihu, Campfire.fm) and reveals key implications for future CQA
system design. We believe this is a first step towards understanding the economy of community-based
knowledge sharing.

\section{Related Work}

\para{Community Based Question Answering (CQA).}
In recent years, researchers have studied CQA services from various
aspects~\cite{Srba:2016:CSC:}. Early studies have focused on identifying domain experts in a CQA community~\cite{PalCK12,Hanrahan:2012:} and routing user questions to the
right experts~\cite{li2010routing,PalWZNS13}. Other works focused on assessing
the quality of existing questions and
answers~\cite{RaviPRK14,Yao:2014,shah2010evaluating,TianZL13,adamic2008knowledge,harper2008predictors,suqi2007} and detecting low quality (or even abusive)
content~\cite{Kayes:2015}. Finally, researchers also studied Q\&A activities in online social networks~\cite{Nichols:2013:,Gray:2013:WKQ}. 
As the sizes of CQA systems rapidly grow, it becomes challenging to engage with experts for timely and high-quality answers~\cite{srba2015stack}.

\para{User Motivations in CQA.} 
A successful CAQ system requires active and sustainable user
participations. Prior works have summarized three main user motivations to
answer questions online: ``intrinsic'',  ``social'' and ``extrinsic''~\cite{jin2013users}. Intrinsic motivation refers to
the psychic reward ({\em e.g.}, enjoyment) that users gain through
helping others~\cite{Yu:2007:,nam2009questions}. Social factors refer
to the benefits of social interactions, {\em e.g.}, gaining respect and
enhancing reputation. Intrinsic and social factors are critical
incentives for non-payment based CQA services~\cite{jin2013users}. 
Extrinsic factors refer to money and virtual rewards ({\em e.g.}, badges and
credit points)~\cite{nam2009questions, grant2013encouraging}.


Monetary incentive is an extrinsic factor implemented in earlier
payment-based CQA services such as Google Answers, Mahalo, ChaCha and
Jisiklog~\cite{chen2010knowledge,hsieh2010pay,lee2013analyzing,lee2012understanding}. Most
of them are defunct. Compared with Fenda and Whale, these systems focus
on building a CQA market on an anonymous crowd, instead of a social
network that engages real-world experts. Users are primarily driven
by financial incentives without a strong sense of
community~\cite{lee2013analyzing,hsieh2009mimir}. This is concerning
since research shows monetary incentive plays an important role in
getting users started in CQA, but it is the social factors that
contribute to the persistent participation~\cite{raban2008incentive}. 


Researchers have studied the impact of monetary incentives but the conclusions vary. Some researchers find
that monetary incentives improves the answer
quality~\cite{harper2008predictors} and response rate in social Q\&A~\cite{Zhu:2016:MYS:}. Others suggest that payments merely reduce the response delay but have no significant impact on the answer quality~\cite{chen2010knowledge,jeon2010re,hsieh2010pay}. 
Studies also show that payment-based Q\&A can reduce low-quality
questions since users are more selective regarding what to ask~\cite{hsieh2009mimir,hsieh2010pay}. 



\para{Crowdsourcing Marketplace.}
Broadly speaking, a payment-based CQA service is a specialized crowdsourcing
marketplace. Today, most crowdsourcing marketplaces ({\em e.g.} Mechanical Turk) are
``customer-driven'' where customers post their tasks and set the task
price~\cite{userstudies-chi08}. In such marketplaces, pricing strategy
can affect the work quality and response time~\cite{katmada2016incentive,mason2010financial,DBLP:journals/corr/JoglekarGP14a}. 
Fenda and Whale, on the other hand, represent ``supplier-driven'' marketplaces where
experts (the suppliers) get to set the price for their answers
(products). Part of our analysis is to understand users' pricing strategies and its impact on their Q\&A
activities and financial income.

\section{Research Questions and Method}
Systems like Fenda and Whale are leading the way to socially engage
with real-world experts for question answering. The introduction of
monetary incentives makes user interactions even more
complex. If not carefully designed, monetary incentives can lead the
systems down to the wrong path with users chasing financial profits
and losing engagement in the long run. In this paper, we use Fenda as
the primary platform to investigate how monetary incentives impact the user
behavior and engagement-level in CQA systems. We also include Whale (a
younger and smaller system) in our analysis for comparison and
validation purposes.

We choose Fenda and Whale for two main reasons. First, Fenda and Whale
represent the first supplier-driven CQA marketplaces with a unique
incentive model to motivate both question askers and
respondents. Second, the system (Fenda in particular) has received an initial success with a significant volume of data and
revenue flow. We aim to understand the reasons behind their success and potential
problems moving forward, which will benefit future CQA design. We did not include Quora
since Quora hasn't accumulated sufficient paid content ({\em e.g.},
$<$ 10 paid answers per month).


\para{Background of Fenda.}
Fenda is a payment-based Q\&A app in China, which
connects users in a Twitter-like social network. Launched in May 2016, Fenda quickly gained 10 million
registered users and over 2 million US dollars worth of questions
answers in the first two months~\cite{news1-fenda16,news3-fenda16}. 

As shown in Figure~\ref{fig:fenda}, Fenda has a unique monetary incentive model to reward both
question askers and answerers. A user (asker) can ask another user (answerer) a question by paying the
price set by the answerer. The answerer then responds over the phone
by recording a 1-minute audio message. If the answerer doesn't respond
within 48 hours, the payment will be refunded. Any other user on Fenda
can listen to the answer by paying a fixed amount of 1 Chinese Yuan
(\$0.14), and it will be split evenly between the asker and
answerer. A good question may attract enough listeners to compensate
the initial cost for the asker. Users set the price for their
answers and can change the price anytime. Fenda charges
10\% of the money made by any user.  
 
There are two types of users on Fenda: verified real-world experts ({\em e.g.},
doctors, entrepreneurs, movie stars) and normal users. There is an
{\em expert list} that contains all the experts that have been
verified and categorized by the Fenda administrators. Users can browse questions from the
social news feed or from the public stream of popular answers (a small
sample). To promote user engagement, Fenda selects 2-4
answers daily on the public stream for free-listening for a limited
time. 
 
\para{Background of Whale.}
Whale is a highly similar system launched in the US in September 2016. 
There are a few differences: First, Whale
users record video (instead of audio) as their
answers. Second, Whale has free questions and paid questions. For paid
questions, it is also the answerer that sets the price, but Whale takes a
higher cut (20\%) from the question fee. Third, listeners use the
virtual currency ``whale coins'' to watch the paid answers. Users can
receive a few {\em free coins} from the platform after
logging-in each day, and they may also purchase {\em paid coins} in
bulks (\$0.29 -- \$0.32 per coin). Only when a listener uses {\em paid} coins to unlock a
question will the asker and answerer receive the extra payment (\$0.099 each).

\para{Our Questions.}
In the following, we use Fenda and Whale as the platform to analyze
how monetary incentives impact user behavior and their
engagement-level. We take a data-driven approach to answer the
following key questions. 

\begin{packed_itemize}
\item First, as an expert-driven CQA system, to what extent does the
  system rely on experts to generate content and revenue? What
  types of experts are more likely to make a profit? 

\item Second, how does the monetary incentive affect the question
  answering process? Does money truly enable on-demand question
  answering from experts? Can users make money by asking (good)
  questions? Will monetary incentives encourage users to game the
  system for profits? 

\item Third, in this supplier-driven market, how do users set and
  dynamically adjust the price of their answers? How does the pricing
  strategy affect their income and engagement-level over time? 

\end{packed_itemize}

\section{Data Collection}
\label{sec:data}
We start by collecting a large dataset from Fenda and Whale through their
mobile APIs. Our data collection focused on user profiles, which
contained a full list of historical questions answered by the
user. To obtain a large set of active users, we explored different
options (some of which did not work). First, we find that there is no
centralized list to crawl all registered users. Second, a
user's follower list is not public (only the total number is visible),
and thus social graph crawling is not feasible. To these ends, we
started our crawling from the expert list. For each expert, we
collected their answered questions and the askers of those
questions. Then we collected the askers' profiles to get their
answered question list and extract new askers. We repeated this
process until no new users appeared. In this way, we collected a large
set of active users who asked or answered at least one question.\footnote{Our
  study has received IRB approval: protocol \#
  16-1143.}

We collected data from Fenda in July 2016.  The dataset contains
88,540 user profiles and 212,082 question-answer pairs ranging from
May 12 to July 27, 2016. Each question is characterized by the asker's userID, question text, a
timestamp, question price, and the number of listeners. Each answer is
characterized by the answerer's userID, a length of the audio and a
timestamp. UserIDs in our dataset have been fully anonymized.
We briefly estimated the coverage of the Fenda dataset. Fenda announced that they had 500,000 answers as of June 27,
2016~\cite{news1-fenda16}. Up to the same date, our dataset covers
155,716 answers (about 31\%). For Whale, we collected 1,419
user profiles and 9,199 question-answer pairs (1114 paid questions and
8085 free questions) from September 7, 2016 to March 8, 2017. 
It is difficult to estimate the coverage of the Whale dataset since there
is no public statistics on the Whale user population.  
Table~\ref{tab:data_stat} shows a summary of our data. 



\begin{table}
\begin{small}
\centering
\scalebox{0.952}{
\begin{tabular}{|c|c|c|c|c|c|}
\hline
Service & Date & \# of &\# of &\# of &\# of\\
Name & (2016-2017)& Questions & Users &  Askers & Answerers \\
\hline
Fenda&5/12/16 -- 7/27/16 & 212,082 & 88,540 & 85,510 & 15,529\\
\hline
Whale&9/07/16 -- 3/08/17 & 9,199 & 1,419 & 1,371 & 656\\
\hline
\end{tabular}}
\caption{Summary of Fenda and Whale dataset.}
\label{tab:data_stat}
\vspace{-0.18in}
\end{small}
\end{table}

\begin{figure*}
\begin{center}
\begin{minipage}[t]{0.24\textwidth}
\includegraphics[width=0.99\textwidth]{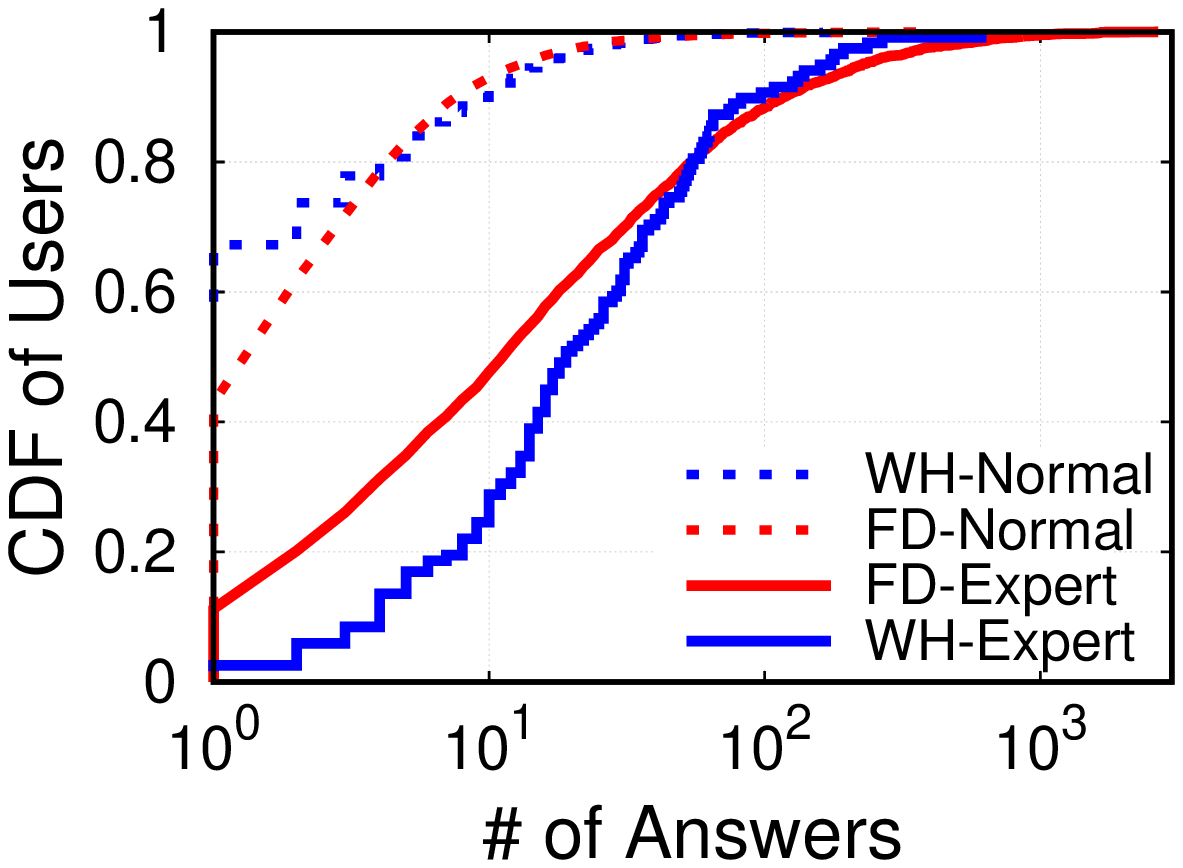}
\vspace{-0.1in}
\caption{\# of Answers per answerer.}
\label{fig:user_answer}
\vspace{-0.1in}
\end{minipage}
\hfill
\begin{minipage}[t]{0.24\textwidth}
\includegraphics[width=0.99\textwidth]{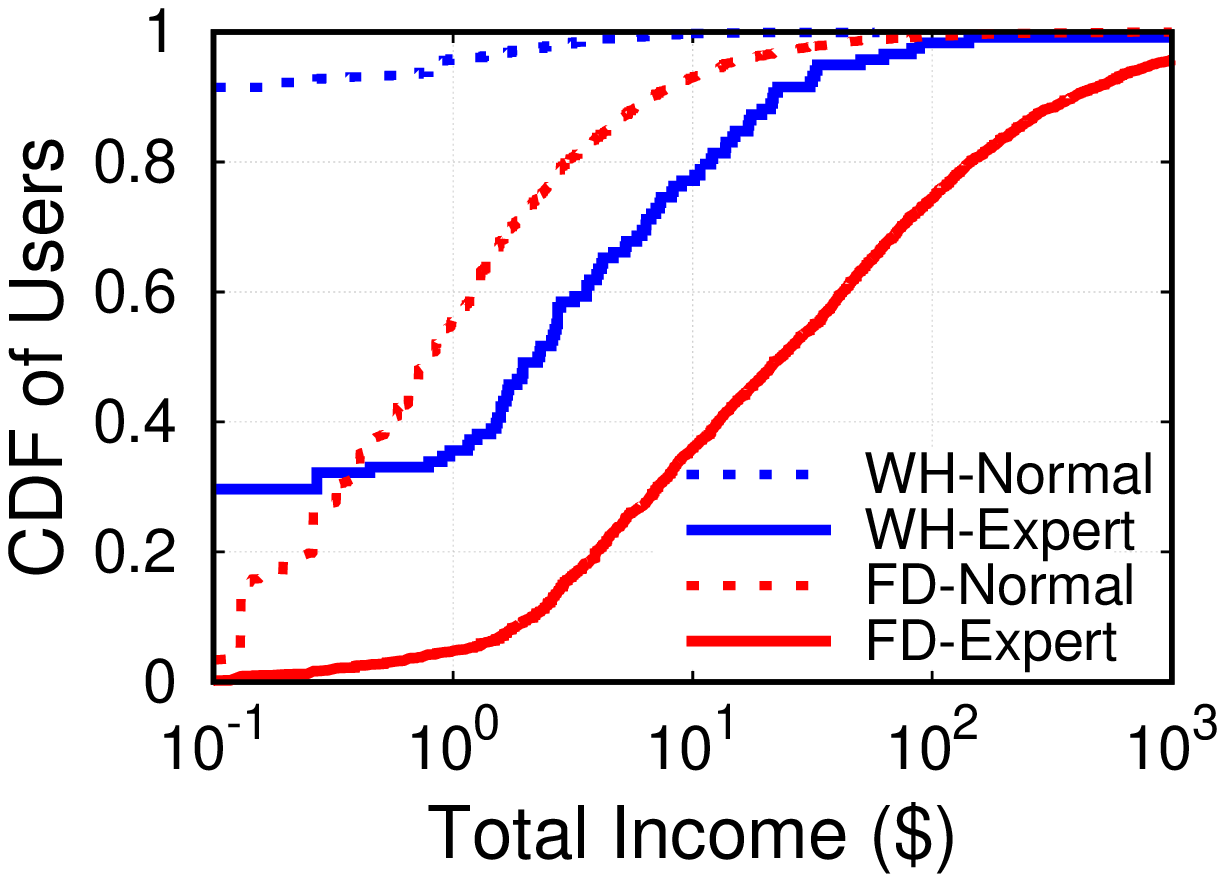}
\vspace{-0.1in}
\caption{Total income per answerer.}
\label{fig:user_income}
\vspace{-0.1in}
\end{minipage}
\hfill
\begin{minipage}[t]{0.24\textwidth}
\includegraphics[width=0.99\textwidth]{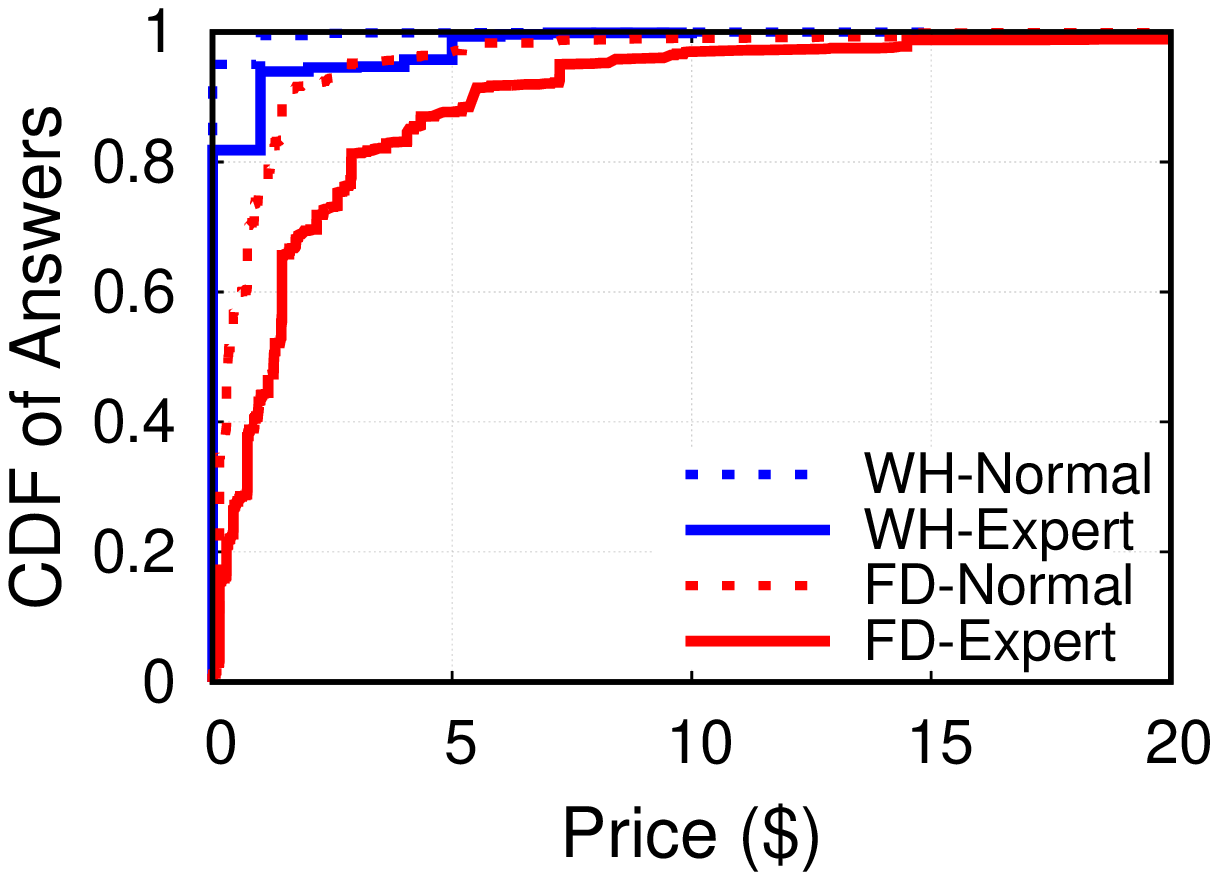}
\vspace{-0.1in}
\caption{Price of each answer.}
\label{fig:answer_price}
\vspace{-0.1in}
\end{minipage}
\hfill
\begin{minipage}[t]{0.24\textwidth}
\includegraphics[width=0.99\textwidth]{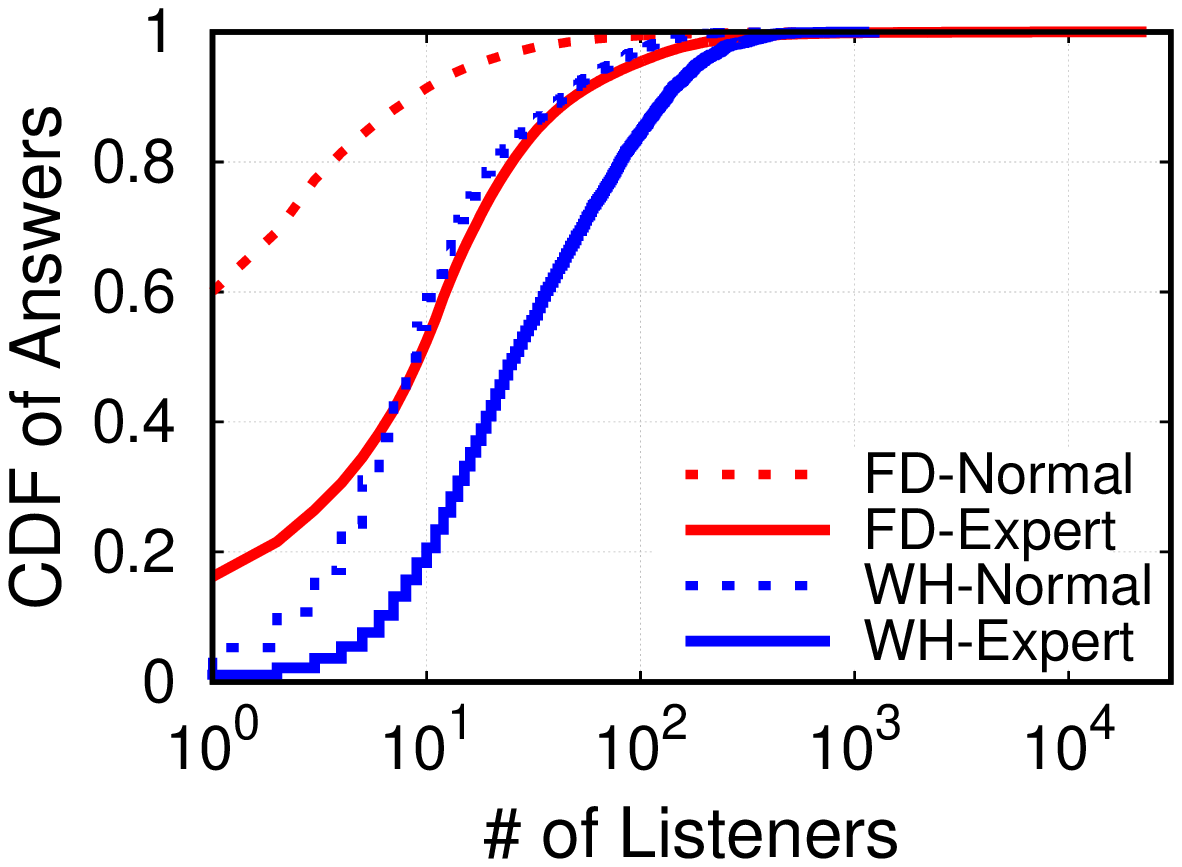}
\vspace{-0.1in}
\caption{\# of Listeners per answer.}
\label{fig:answer_listener}
\vspace{-0.1in}
\end{minipage}
\end{center}
\end{figure*}

\section{Engaging with Domain Experts}
As a CQA system driven by real-world experts, we first explore the
roles and impact of domain experts in the system. More specifically,
we examine the contributions of domain experts to the
community in terms of generating content and driving financial
revenue.



 





Fenda maintains a list of verified experts and celebrities, who are
typically already well-known in their respective
domains. As of the time of data collection, there were 4370 verified
experts classified into 44 categories by Fenda administrators. We refer these 4,370
users as  {\em experts} and the rest 84,170 users as {\em normal
  users}.  Whale has a similar expert list (118 experts), and we refer
the rest 1301 Whale users as normal users. 





\subsection{Question Answering}
The small group of experts have contributed to a
significant portion of the answers. Out of the 212K answers in the Fenda
dataset, 171K (81\%) are from experts. Using this dataset, we can
briefly estimate the experts' contribution in the context of the entire
network. On June 27 2016, Fenda officially announced total 500K
answers and 10 million users~\cite{news1-fenda16}. Up to the same
date, our dataset shows the 4,370 experts (0.44\% of the population)
have contributed 122K answers (24.4\% of total answers). Individually,
experts in Fenda (FD) also answered significantly more questions than normal users
as shown in Figure~\ref{fig:user_answer}. Whale (WH) has a similar
situation where 118 experts (8\% of users) have contributed 4,967 answers (54\%
of answers).





\subsection{Money}
Experts play an important role in driving revenue. In total,
the questions in the Fenda dataset were worth \$1,169,994
including payments from askers and listeners.\footnote{We convert
  Chinese Yuan to US dollar based on \$1 = 6.9
  Yuan.}  Experts' answers generated \$1,106,561, counting for a
dominating 95\% of total revenue in our dataset. To gauge experts'
contribution in the context of the entire network, we again performed
an estimation: Fenda reached 2 million revenue as of June
27 in 2016~\cite{news1-fenda16}. Up to this same date (June 27), expert
answers in our dataset have attracted \$909,876, counting for a
significant 45\% of the 2 million
revenue. Figure~\ref{fig:answer_price} and
Figure~\ref{fig:answer_listener} show that, on average, experts charge higher (\$2.9 vs. \$1.0)
and draw more listeners (27 vs. 5) than normal users.

Individually, experts also make more money than normal
users. Figure~\ref{fig:user_income} shows the total income for users
who answered at least one question. On Fenda, 50\% of experts made more than \$23, while a
small group of experts (5\%) made more than \$1000. The highest
earning is \$33,130 by Sicong Wang, a businessman and the son of a
Chinese billionaire. He answered 31 questions related to gossip and
investment. He charged \$500 for each of his answers, which drew 9484
listeners (\$664 extra earning) per answer on average.  

On Whale, experts are also the major contributors to the
revenue flow. The total collected questions on Whale worth \$2,309 and
experts contributed to \$2,028 (89\%). Compared
with Fenda (FD), Whale (WH) users earned significantly less money (Figure~\ref{fig:user_income}). A possible reason, as shown
in Figure~\ref{fig:answer_price}, is that most users (more than 80\%)
provide answers for free.

\begin{table}[t]
\begin{small}
  \centering
  \scalebox{0.91}{
  \begin{tabular}{|ccc|ccc|}
    \hline
 \multicolumn{3}{|c}{Fenda} & \multicolumn{3}{|c|}{Whale} \\ \hline
Category & Tot. Income& Experts  & Category & Tot. Income   & Experts       \\ \hline
 Health & \$123K (12\%) & 204 & Startups & \$1.9K (72\%) & 63  	\\ \hline
Career & \$81K (8\%)  & 222 & Tech & \$1.8K (72\%)& 61  	\\ \hline
Business & \$81K (8\%) & 108 & Entertain. & \$877 (33\%)  & 2 	\\ \hline
Relation. & \$73K (7\%)   & 90 & Snapchat & \$869 (33\%)  & 1   \\\hline
Movies & \$52K (5\%)& 84& Motorcycle & \$869 (33\%)  & 1 	\\ \hline  
Entertain. & \$52K (5\%) & 51 & Marketing& \$471 (18\%)  & 20 	\\ \hline  
Academia & \$49K (5\%) & 64 & Design & \$383 (14\%)  & 15 	\\ \hline 
Media & \$45K (5\%) & 138& Travel & \$203 (8\%)  & 18  	\\ \hline
Real Estate & \$43K (4\%) & 28 & Fitness& \$191 (7\%)  & 19 	\\ \hline
Education & \$39K (4\%)  & 174 & Finance & \$141 (5\%)  & 8	\\ \hline
 \end{tabular}
  }
  \caption{Top 10 expert categories with highest total earnings.}
  \label{tab:top_income_category}
  \vspace{-0.12in}
 \end{small}
\end{table}


\subsection{Expert Categories}
Experts of different categories have distinct earning
patterns. Table~\ref{tab:top_income_category} shows the top 10
categories ranked by the total earnings per category. 
In Fenda, the most popular experts are related to professional
consulting. The top category is {\tt health},  followed by {\tt career},
{\tt business}, and {\tt relationship}.  In the {\tt health} category,
many experts are real-world physicians and pediatricians. They give
Fenda users medical advice on various (non-life-threatening) issues such as
headache and flu with the expense of several dollars. Other popular categories such as
{\tt movies} and {\tt entertainments} contain questions to
celebrities about their insider experience, gossip and opinions on
trending events. 

Whale, on the other hand, has fewer experts. The
highest earning experts are related to {\tt startups} and {\tt
  technology}. Note that Whale experts can belong to multiple categories, so
the percentage of accumulated income exceeds 100\%.   

\begin{figure}
\begin{center}
\begin{minipage}[t]{0.49\textwidth}
\includegraphics[width=0.99\textwidth]
 {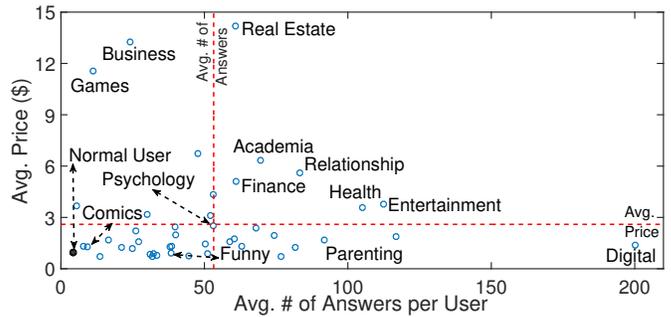}
 \caption{Scatter plot of the average number of answers per expert and
   average price in each expert category (Fenda). The red lines represent the
   corresponding average values across all experts.}
\label{fig:category_income}
\vspace{-0.1in}
\end{minipage}
\end{center}
\end{figure}

In Figure~\ref{fig:category_income}, we further illustrate the
distinct earning patterns of Fenda experts. We omit the result for
Whale due to its short expert list. For each category, we compute the
average price and the number of answers per expert. The red lines
represent the average values across all experts, which divide them
into 4 sections.  Experts in {\tt health}, {\tt entertainment},
{\tt relationship}, and {\tt real estate} often charge high and answer
many questions. These experts are among the top-earning groups; Experts in {\tt business} set
the price high but don't answer many questions; Less-serious
categories such as {\tt funny} and {\tt comics} have fewer and less
expensive questions. Finally, {\tt digital} represents experts 
who answer lots of questions with a low price. The results also
reflect the different user perceived values for different domain knowledge.

\begin{figure*}
\centering
\begin{minipage}{0.32\textwidth}
\includegraphics[width=1\textwidth]
{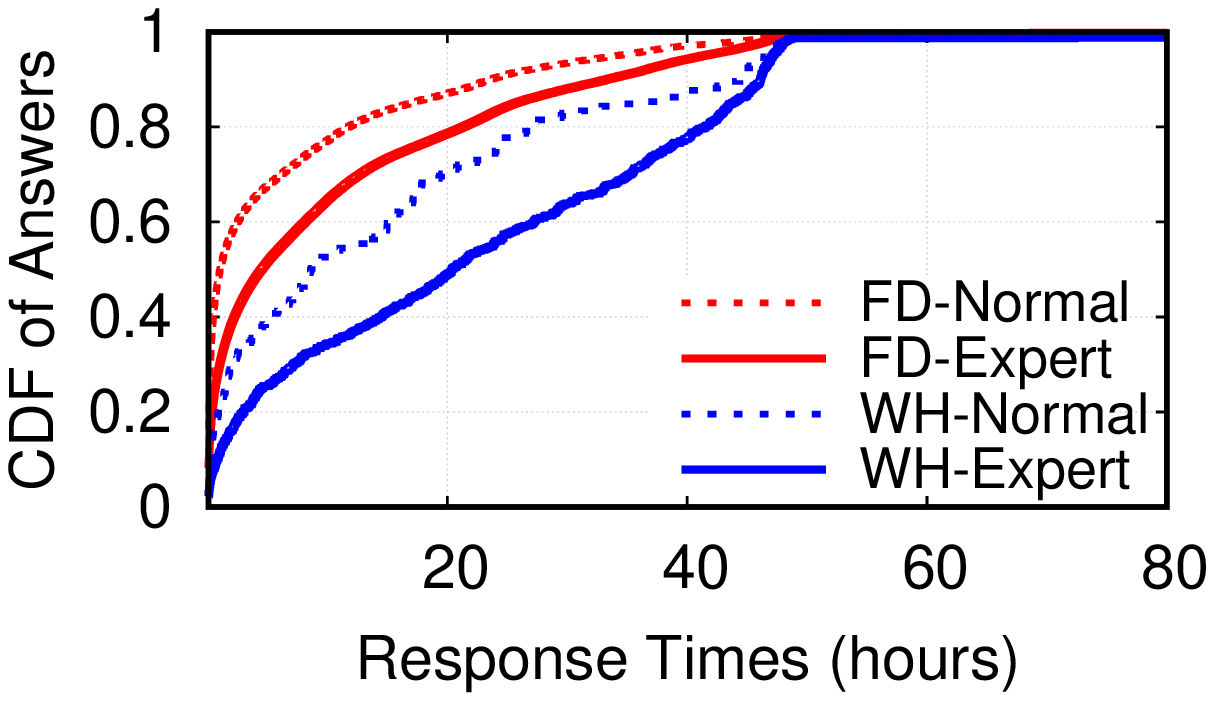}
\vspace{-0.12in}
\caption{Response time of answers.}
\label{fig:answer_time}
\vspace{-0.1in}
\end{minipage}
\hfill
\begin{minipage}{0.32\textwidth}
\includegraphics[width=1\textwidth]
{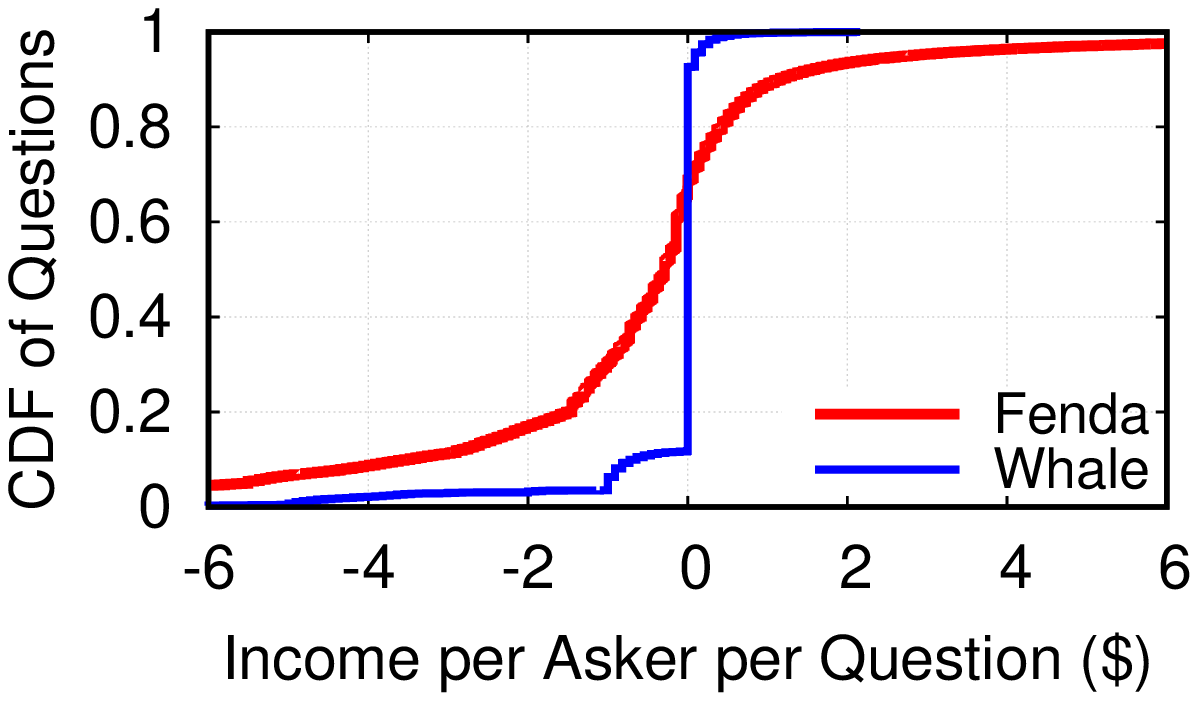}
\vspace{-0.12in}
\caption{Income of askers per question. }
\label{fig:questionincome}
\vspace{-0.1in}
\end{minipage}
\hfill
\begin{minipage}{0.32\textwidth}
\includegraphics[width=1\textwidth]
{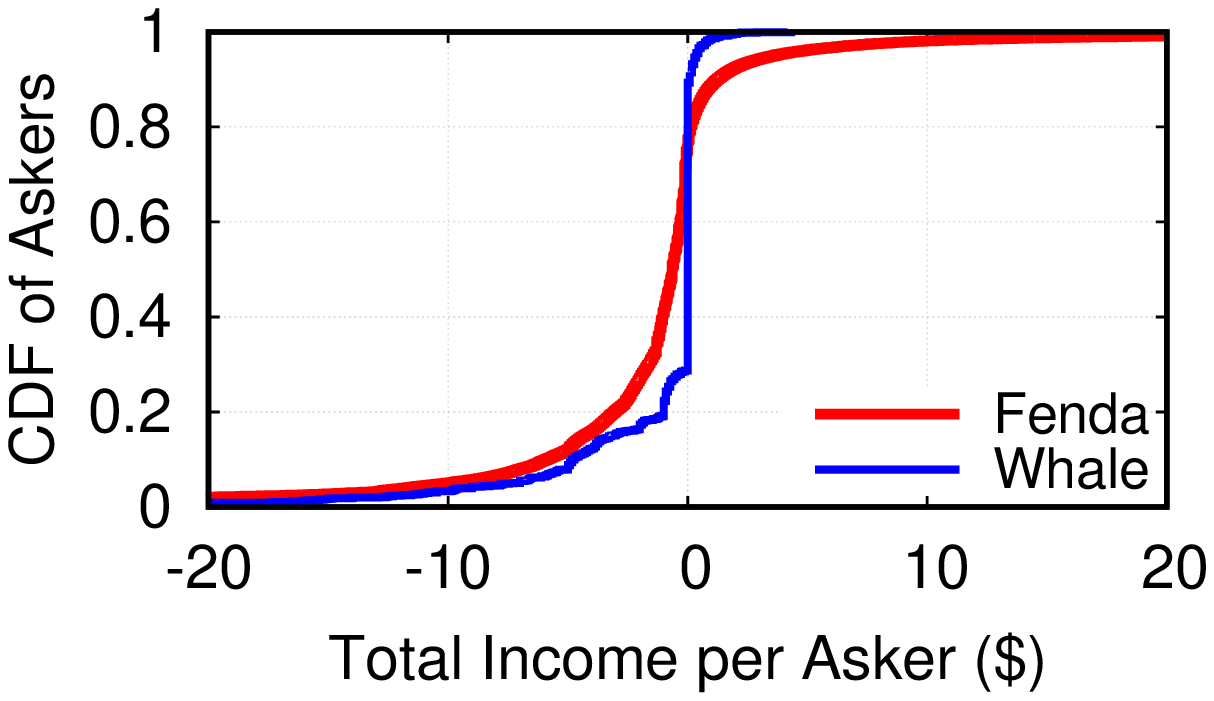}
\vspace{-0.12in}
\caption{Total income of askers.}
\label{fig:incomedistribution}
\vspace{-0.1in}
\end{minipage}
\end{figure*}

\section{Impact of Monetary Incentives}
So far we show that Fenda and Whale are highly dependent on domain
experts' contribution. Then the question is how to motivate experts to
deliver timely and high-quality answers. In this section, we perform
extensive analysis on the monetary incentive model to understand its
impact on user behavior. Noticeably, Fenda and Whale use money to
reward both question answerers and askers. To this end, we first
analyze {\em answerers} to understand how they price their
answers, and whether payments lead to
on-demand responses. Second, we focus on {\em askers} analyzing
whether and how users make money by asking the right
questions. Finally, we seek to
identify {\em abnormal users} such as ``bounty hunters'' who aggressively or strategically
game the system for profits.  


    


  \begin{table}
  \centering
  \begin{small}
        \begin{tabular}{|r||c|c|}
        \hline
  Behavior Metric &  Fenda & Whale  \\  \hline
  \# Followers &  0.53* &  0.30*       \\ \hline
  Avg. \# Listeners  & 0.65*  & 0.08*      \\ \hline
  \# Questions Answered &  0.04* &  0.14* \\ \hline
  Avg. Response Time &  0.01 &  -0.07     \\ \hline
        \end{tabular}
                \caption{Pearson correlation between a user's answer price and key behavior metrics. * indicates significant correlation with  $p<0.05$. }
 \label{tbl:whale_CorrelationCoefficient}
\vspace{-0.07in}
        \end{small}
    \end{table}

\subsection{Answerers}
To motivate users (particularly domain experts), both Fenda and Whale
allow users to determine the price for their answers. In the following, we investigate how money affects the way
users answer questions. Particularly, we examine if monetary
incentives truly enable on-demand quick answers. 


\para{Setting the Answer Price.} To understand how users set a price for
their answers, we calculate the Pearson correlation~\cite{Sheskin:2007:HPN:1529939} between a
user's price and different behavior metrics. In
Table~\ref{tbl:whale_CorrelationCoefficient}, we observe that the
price has positive and significant correlations with the number of
followers, listeners, and answered questions. A possible explanation is
that users with many followers and listeners are real-world
celebrities who have the confidence to set a higher price. The higher
price may also motivate them to answer more questions. Note that these
are correlation results, which do not reflect causality. 


Surprisingly, there is no significant correlation between price and
response time (for both Fenda and Whale). This is different from existing results on
customer-driven CQA markets, where an asker can use a higher payment to collect answers more
quickly~\cite{katmada2016incentive,mason2010financial,hsieh2010pay}. 

\begin{table}
{\centering
\begin{small}
        \begin{tabular}{|r|c|c|l|}
        \hline
Service Name & Avg. Resp.  & Payment  &  Crowdsourcing     \\ 
   & Time (hr) &  Based?   &   or Targeted?     \\ \hline
 Yahoo Answers   & 8.25  &  N & Crowdsourcing  \\ \hline
Fenda   & 10.4 & Y & Targeted   \\ \hline
Whale   & 23.6  & Y & Targeted   \\ \hline
Google Answers   & 36.9  & Y  & Crowdsourcing   \\ \hline
Stack Overflow  & 58.0 & N & Crowdsourcing   \\ \hline

        \end{tabular}
        \caption{Average response time of the first answer (in hours). We compare Fenda and Whale with different CQA sites including  Yahoo Answers~\protect\cite{iconference2014}, Google Answers~\protect\cite{edelman_2011} and StackOverflow~\protect\cite{Mamykina:2011}. }
        \label{tbl:respondtime}
  \end{small}
\vspace{-0.13in}
  }
  \end{table}

\para{Answering On-demand? }
We further examine the response time to see if monetary incentives
truly enable answering questions on-demand. As shown in
Figure~\ref{fig:answer_time}, answers arrive fast on Fenda: 33\% of
answers arrived within an hour and 85\% arrived within a day. Note
that there is a clear cut-off at 48 hours. This is the time when
un-answered questions will be refunded, which motivates users to
answer questions quickly. After 48 hours, users can still answer those
questions for free. We find that only 0.7\% of the answers arrived
after the deadline, but we cannot estimate how many questions remain
unanswered due to the lack of related data. Despite the high
price charged by experts, experts respond slower than normal users. 

The result for Whale is very similar. Figure~\ref{fig:answer_time}
shows that for {\em paid} questions, 50\%--70\% of answers arrived within a
day and normal users respond faster than experts. Comparing to Fenda,
Whale has a slightly longer delay possibly because recording a
video incurs more overhead than recording a voice message. 



We then compare Fenda and Whale with other CQA systems in
Table~\ref{tbl:respondtime}. The response delay in Fenda and Whale is
shorter than that of Google Answers and
StackOverflow, but longer than that of Yahoo Answers. 
As payment-based systems, Fenda/Whale beats Google Answers probably
because Fenda/Whale only asks for a short audio/video, while Google
Answers require lengthy text. Compared to Yahoo Answers, we believe it
is the crowdsourcing factor that plays the role. Systems like Yahoo
Answers crowdsource questions to a whole community where anyone could
deliver the answer. Instead, Fenda/Whale's question is targeted to a
specific user, which may lead to a longer delay even with
payments. 

\subsection{Askers}
Fenda and Whale implement the first monetary incentive model to reward
users for asking good questions. More specifically, once a user's
question gets answered, this user (the question asker) can earn a small
amount of money from people who want to listen to the answer. This model, if executed as designed, should motivate users to contribute
high-quality questions for the community.

\para{Can Askers Make Money? } For each question, the question asker's income is half of listeners'
payments, with Fenda's commission fee and initial question fee
deducted. Our result shows that Fenda askers are motivated to ask good
questions that attract broad interests. As shown in
Figure~\ref{fig:questionincome}, out of all questions, 40\% have
successfully attracted enough listeners to return a positive profit to
the asker. For individual askers, Figure~\ref{fig:incomedistribution}
shows 40\% of them have a positive total income. However, for Whale,
the vast majority of askers did not earn money. Part of the reason is
most people only ask free questions. More importantly, Whale gives
away free coins every day to motivate users to login. If a listener uses free coins
(instead of paid coins), the asker will not receive any money.


\para{How Do Askers Make Money? }
To understand why certain users make money (and others don't), we
compare askers who have positive income with those with negative
income in Table~\ref{tbl:top4earnornot}. Specifically, we examine to
whom they ask questions ({\em i.e.}, the number followers and
listeners of the answerer), average question price, and total
questions asked. A two-sample t-Test~\cite{Sheskin:2007:HPN:1529939}
shows the significance of the differences between the two groups of askers. 

On Fenda, users of positive income are more likely send questions to
people who have more listeners and charge less. The counter-intuitive
result is the {\em number of followers}: asking people with more
followers is more likely to lose money. A possible explanation is the
inherent correlation between a user's number of followers and her
answer price --- famous people would charge higher and the money from
listeners cannot cover the initial cost. We also observe that askers
with a higher income often asked more questions. Again, correlation
does not reflect causality: it is possible that the positive income
motivates users to ask more questions, or people who asked more
questions get more experienced in earning money.

It is hard to interpret the Whale results in
Table~\ref{tbl:top4earnornot} since only a very small of fraction of
askers have a positive income (Figure~\ref{fig:incomedistribution}). Noticeably, askers with positive
income exclusively ask free questions (average price = 0).

\begin{table}
\centering
        \begin{small}
        \begin{tabular}{|r||c|c|c|c|c|c|}
        \hline
      & \multicolumn{3}{c|}{Fenda} &
     \multicolumn{3}{c|}{Whale}  \\ 
\cline{2-7}
Behavior & Askers  & Askers &  $p$ &  Askers  & Askers &  $p$ \\ 
Metric  & $\$>$0 & $\$\leq$0 &  & $\$>$0 & $\$\leq$0 &  \\ \hline
Avg. Followers  & 2155.5  & 3758.5 &* & 750.2 &790.0 &  \\ \hline
Avg. Listeners  & 55.2  & 16.9 &* &28.3  & 38.1 &*  \\ \hline
Avg. Price  & 1.58  & 4.58 &* & 0.0 & 0.3 &*   \\ \hline
Avg. Questions  & 3.99  & 1.86 &* & 5.4  & 6.6 &   \\ \hline
        \end{tabular}
    \caption{Two sample t-test compares the behavior metrics for
      askers with positive income and those with negative
      income. * indicates the differences between the two types of
      askers are significant with $p<0.05$. }
      \label{tbl:top4earnornot}
        \end{small}
\vspace{-0.12in}
    \end{table}


 


\begin{figure}
\begin{center}
\begin{minipage}[t]{0.49\textwidth}
\includegraphics[width=0.99\textwidth]
 {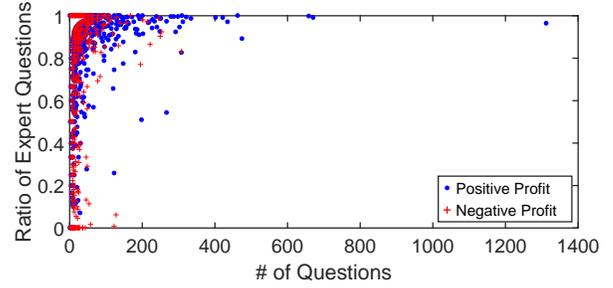}
 \caption{Total \# of questions of each asker vs. the ratio of
   questions to experts in Fenda. Blue dots (red crosses) represent
   askers with positive (negative) total income. The figure is better
   viewed in color.}
\label{fig:expert_question_ratio}
\end{minipage}
\vspace{-0.15in}
\end{center}
\end{figure}



\subsection{Abnormal Users}
Next, we examine suspicious users in the CQA market who seek to game
the system for financial profits.

\para{Bounty Hunters.}
We first focus on askers who
aggressively ask questions to gain profits (or ``bounty hunters''). Our intuition is that those
users would ask a lot of questions, particular to experts. To identify potential bounty hunters in Fenda, we examine outliers in
Figure~\ref{fig:expert_question_ratio}, which is a scatter plot for
the number of questions a user asked versus the ratio of questions to
experts. We find clear outliers at the right side ({\em e.g.}, users
with $>$100 questions). Most of these users end up with positive
income. They asked way more questions than other users (who asked
2.27 questions on average), and exclusively interact with experts (ratio of expert
questions is close to 1). The most extreme example is a user who asked
more than 1300 questions in two months, with 95\% of questions to
experts. This user earned \$194.20, which is much higher than the average
income of askers (-\$1.95). 

To further examine these outliers, we select askers who asked more
than 100 questions. This gives us 111 users who count for 0.13\% of
askers in our dataset but aggressively asked 11\% of the questions. In
addition, they seem to carefully target experts who charge a lower
price  (\$0.80 per answer) but still draw significant
listeners (15.5 per answer). As a comparison, the rest of
the experts on average charge \$2.49 and draw 23.0 listeners per answer. 

We performed the same analysis on Whale and
did not find such outlier users because most askers on Whale did not
make a positive profit (Figure~\ref{fig:incomedistribution}). We omit
the results for brevity. 


\begin{figure}
\begin{center}
 \begin{minipage}[t]{0.233\textwidth}
 \subfigure[Fenda]{
\includegraphics[width=0.99\textwidth]
 {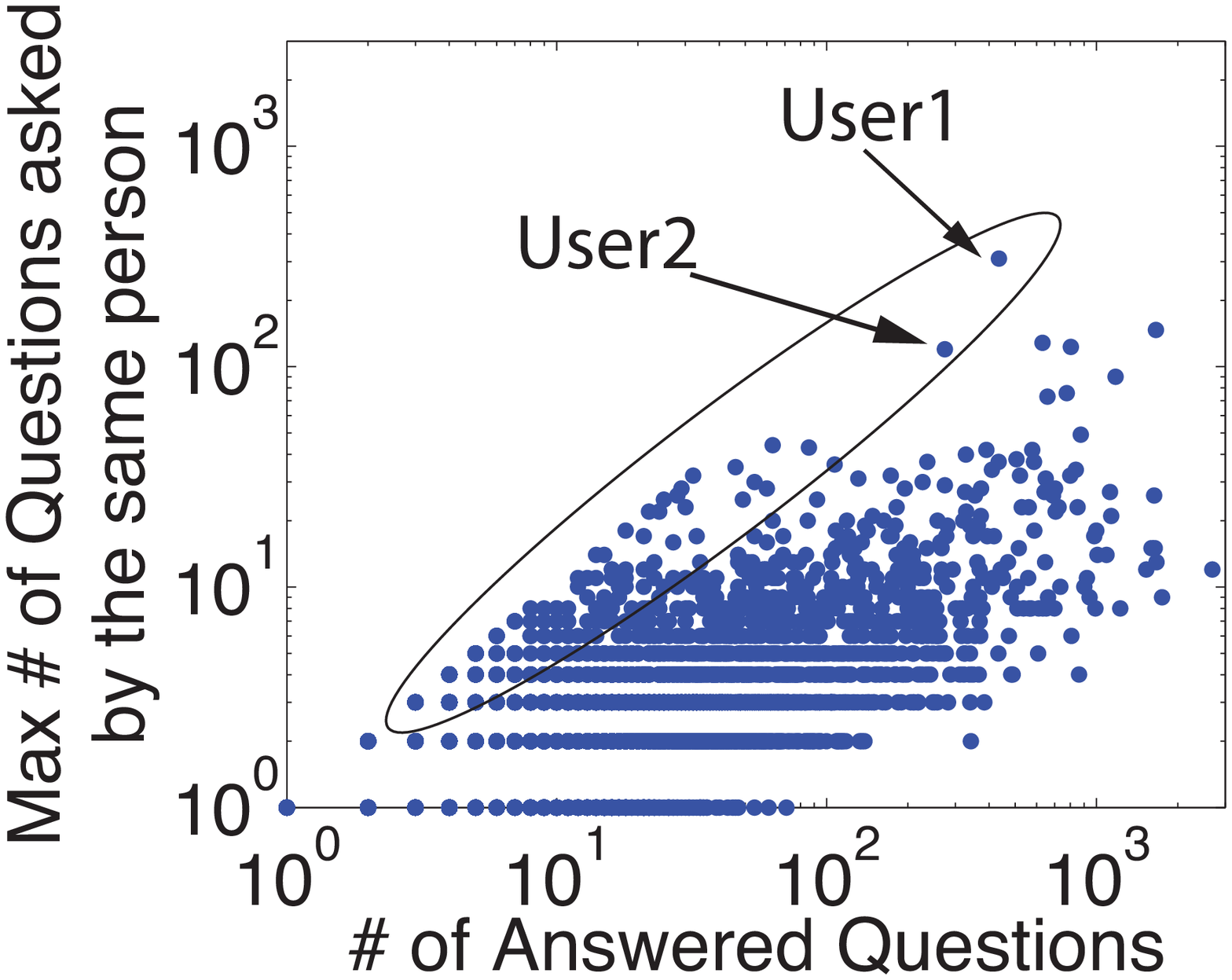}
\label{fig:conspire1}
}
 \end{minipage}
  \begin{minipage}[t]{0.233\textwidth}
   \subfigure[Whale]{
\includegraphics[width=0.99\textwidth]
 {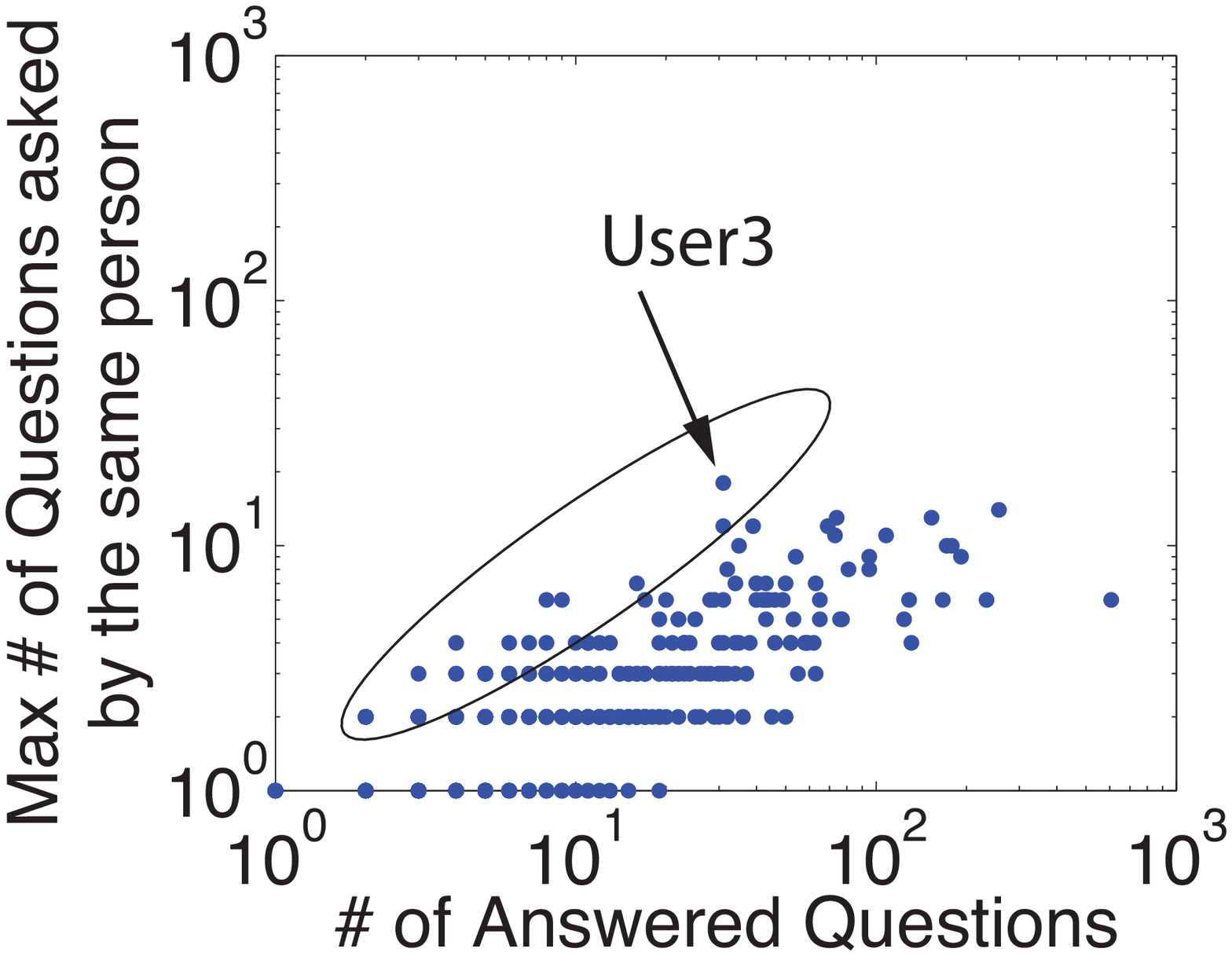}
\label{fig:conspire2}
}
 \end{minipage}
 \caption{Total \# of questions of each answerer vs. Maximum \# of
   questions asked by the same person. Dots in the circled area
   are likely the collusive users. We select three example users for
   case studies. }
\label{fig:conspire}
\end{center}
\vspace{-0.15in}
\end{figure}

\para{Collusive Users.}
In addition to bounty hunters, there could be ``collusive'' users who
work together to make money. For example, an asker may collude with
an answerer by asking many questions (with an extremely low price) to create the
illusion that the answerer is very popular. Then both the asker and the
answerer can make money from the listeners of these questions. 
This collusion may be even conducted by a single attacker who
controlled both the asker and answerer accounts. 

To identify collusive users, we focus on answerers whose questions are
primarily asked by the same user. Figure~\ref{fig:conspire} shows
a scatter plot for the number of questions a user answered versus
the maximum number of these questions asked by the same
person. Users that are close to the diagonal are suspicious. Take the
two marked dots in Figure~\ref{fig:conspire1} for example,
{\em user1} answered 435 questions and 309 (71\%) were asked by the
same asker. We notice that this asker did not ask any other users any questions. The
questions between these two users charge \$0.16 each, which is much
lower than {\em user1}'s other questions (\$0.25 on average). By using a
lower price for collusion, the two users can minimize their loss --- the 10\% commission
fee to Fenda. In this way, {\em user1} earned \$689.9 in total and this asker also
earned \$244 from the listeners. The second example follows the same
pattern. 120 out of 274 questions that {\em user2} answered were from
the same asker (who only asked 128 questions in total). This leads to
financial returns for both {\em user2} (\$116.42) and the asker (\$27.5). 

Figure~\ref{fig:conspire2} shows the result of Whale. The example user (user3) answered 31 questions, 18
of which were from the same asker. This asker only asked these 18
questions and all 18 questions were free of charge. This is likely an attempt to
boost user3's popularity. 


\para{Discussion.}
Our analysis shows that monetary incentives did foster manipulative
behavior. On the positive side, these users (bounty hunters or collusive users) are actually working hard to come
up with interesting questions in order to earn money from listeners. On the negative
side, such behavior has a disruptive impact to the marketplace. 
For example, bounty hunters are injecting a large volume of
questions to experts. 
The large volume of questions act as spam to experts, blocking other users' chance to get the experts' attention. For collusive
behavior, it creates a fake perception of popularity, which could
mislead listeners to making the wrong spending. In addition, collusion
also introduces unfairness to other honest experts, which is likely
to hurt the sustainability of the community in the long run. 

Our analysis focuses on the most likely attacks, and there could be other types of attacks. For
example, answerers and listeners may also collude to bootstrap
the ``listening count'' for a question and lure innocent users to
pay for listening. However, we don't have the related data ({\em e.g.},
listeners' IDs) to analyze this attack.

\section{Dynamic Pricing and User Engagement }
As supplier-driven marketplaces, Fenda and Whale allow users to set
the price for their answers. How users set this price may
affect their financial income and their interaction with
other users. In this section, we turn to the {\em dynamic} aspect to
analyze how users adjust their answer prices over time and how
different pricing strategies affect their engagement level. 
Understanding this question is critical since keeping users
(particularly experts) engaged is the key to building a
sustainable CQA service. 

In the following, we first identify common pricing strategies
by applying unsupervised clustering on users' traces of price change. Then we
analyze the identified clusters
to understand what type of users they represent, and how their engagement-level
changes over time.

\begin{table}
\centering
\begin{small}
  \scalebox{0.97}{
\begin{tabular}{|p{0.1cm}|l|l|}
\hline
id &Feature Name  &  Feature Description    \\ \hline\hline
1 &Price Change Freq. & \# of price change / \# answers   \\\hline
2 &Price Up Freq. & \# price up / \# answers  \\\hline
3 &Price Down Freq. & \# price down / \# answers  \\\hline
4 &Price Up - Down &  (\# price up -  \# price down) / \# answers      \\\hline
5 &Price Up Magnitude   &   Average percentage of price increase  \\\hline
6 &Pirce Down Magnitude   &  Average percentage of price decrease     \\\hline
7 & Consecut. Same Price&  Max \# consecutive same price   / \# answers    \\\hline
8 & Consecut, Price Up &  Max \# consecutive price up / \# answers     \\\hline
9 & Consecut. Price Down &  Max \# consecutive price down / \# answers    \\\hline
\end{tabular}
}
\caption{A list of features for price change dynamics.}
\label{tbl:table-feature}
\vspace{-0.15in}
\end{small}
\end{table}


\subsection{Identifying Distinct Pricing Strategies}
To characterize users' dynamic price change, we construct a list
of features to group users with similar price change patterns. 

\para{Key Features. }
For each user, we model their price change as a sequence of
events. Given user $i$, our dataset contains the complete list of her
answers and the price for each answer. We use $P_i$ to denote user
$i$'s price sequence  $P_i = [p_{i,1}, p_{i,2}, ..., p_{i,N_i}]$ where
$N_i$ is the total number of answers of user $i$.  A price change
event happens when  $p_{i,j-1} \neq  p_{i,j}$ for any $j \in
[2,N_i]$. We denote the price change sequence as $C_i =  [c_{i,1},
c_{i,2}, ... c_{i,M_i}]$ where $M_i$ is a number of times for price
change and $c_{i,j}$ is a price change event (price-up, price-down, or
same-price). 

Table~\ref{tbl:table-feature} list our 9 features: the
overall frequency of price change ({\em i.e.}, $ \frac{M_i}{N_i}$), a
frequency for price-up and price-down, and the frequency
difference between price-up and down.  In addition, we consider the
average price change magnitude for price-up and price-down
events. Finally, we consider the maximum number of consecutive events
of same-price, price-up and price-down in the sequence.  


\para{User Clustering.}
Based on these features, we then cluster similar users into
groups. First, we compute the pair-wise
Euclidean distance between users based on their feature vectors. This
produces a fully connected similarity graph~\cite{chi16} where each
node is a user and edges are weighted by distance. 
Then, we apply hierarchical clustering algorithm
\cite{Fortunato201075} to detect groups of users with similar price
change patterns. We choose hierarchical clustering for two reasons: 1)
It does not pre-define the number of clusters. 2) It is deterministic
and the clustering result does not depend on the initial seeding.

To determine the number of clusters, we use {\em modularity}, a
well-known metric to measure clustering quality~\cite{Fortunato201075}. High modularity means users are more
densely connected within each cluster than to the rest of the users. 
We choose the number of clusters that yields the highest modularity. 

For this analysis, we only consider users who have answered enough
questions (more than 10). Otherwise, discussing their dynamic price change and
engagement would be less meaningful. On Fenda, this filtering produces
2094 users who have answered 171,322 questions (85\% of all questions). On Whale, however, only 68 users meet the criteria. These
users answered 986 paid questions (89\% of all paid questions). We
will primarily focus on Fenda, and also include Whale's results for completeness. 



\begin{figure}
  \centering
  \begin{minipage}[t]{0.48\textwidth}
\subfigure[Fenda]{
\includegraphics[width=0.99\textwidth]
 {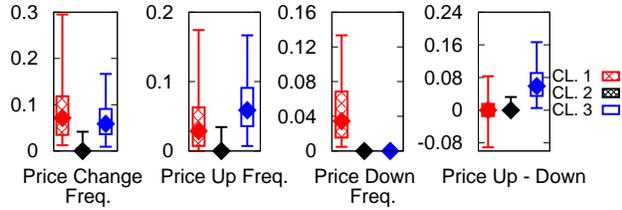}
   \vspace{-0.13in}
   \label{fig:featuredistribution1}}
\subfigure[Whale]{
\includegraphics[width=0.99\textwidth]
 {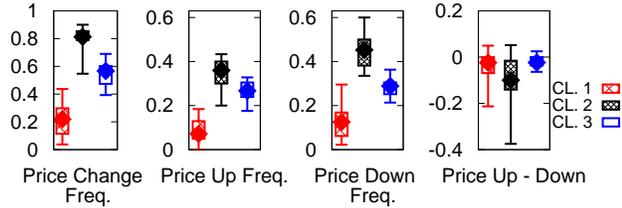}
   \vspace{-0.13in}
   \label{fig:featuredistribution2}}
  \vspace{-0.1in}
   \caption{The distribution of top 4 features for the 3 clusters in
     Fenda and Whale. We depict each distribution with box plot quantiles (5\%, 25\%, 50\%, 75\%, 95\%). The detailed feature
     description is in Table~\ref{tbl:table-feature}.}
   \label{fig:featuredistribution}
\end{minipage}
   \vspace{-0.13in}
\end{figure}


\subsection{Clustering Results}
Our method produces 3 clusters for both Fenda (modularity 0.59) and
Whale (modularity 0.62). To understand the pricing strategy of each
cluster, we plot their feature value distributions in
Figure~\ref{fig:featuredistribution}. Due to space limitation, we plot
4 (out of 9) most distinguishing features that have the largest
variance among the 3 clusters selected by Chi-Squared
statistic~\cite{Sheskin:2007:HPN:1529939}. 
The three clusters on Fenda are: 
\begin{packed_itemize}
\item {\textbf{CL.1 (33\%):} {\em Frequent price up and down.} }
687 users (76\% are experts) who have a high price
  change frequency. Price up and down are
  almost equally frequent. 
\item {\textbf{CL.2 (43\%):} {\em Rarely changing price.}} 908 users
  (76\% are experts) who rarely change their price. 

\item {\textbf{CL.3 (24\%):} {\em Frequent price up.}} 499 users (74\%
  are experts) who increase price frequently but rarely lower their price. 
\end{packed_itemize}

We find that the 3 types of pricing patterns on Fenda correspond to users of
different popularity. As shown in Table~\ref{tbl:overallcompare},
cluster 1 represents the least popular answerers, who have the least
followers and listeners but answered more questions. These users
constantly adjust their price, possibly to test the market. Cluster 3
represents the most popular experts and celebrities. They charge
higher than others and keep increasing the price. Cluster 2 stands
between cluster 1 and 3 in terms of popularity, and its users rarely
change the price.

Whale's 3 clusters only contain 68 users in total. 
We include the results for completeness:  
\begin{packed_itemize}
\item {\textbf{CL.1 (60\%):} {\em Rarely changing price.} } 41 users
  (32\% are experts) with the least
  frequent price change. 

\item {\textbf{CL.2 (22\%):} {\em Frequent price up and down.}}
15 users (87\% are experts) who frequently
change/drop the price. 

\item {\textbf{CL.3 (18\%):} {\em Occasional price up and down.}}
12 users (92\% are experts) who occasionally 
change the price. 
\end{packed_itemize}

As shown in Table~\ref{tbl:overallcompare}, Whale's cluster 3 contains the most
popular users, followed by cluster 1 and 2. Figure~\ref{fig:featuredistribution2} ``Price Up-Down'' shows the
most popular users of cluster 3 are relatively balanced in terms of increasing
versus decreasing the price. The less popular users of cluster 1 and 2 are more
leaning towards decreasing the price. Compared to Fenda, all the clusters of Whale adjust their price rather
frequently. This shows that popular users on Fenda already have the luxury to keep
increasing the price. On Whale, even the most popular users are
frequently adjusting their price, possibly due to the limited earning
opportunities (a much lower payment per question).

\begin{table}[t]
\begin{small}
  \centering
  \begin{tabular}{|c|ccc|ccc|}
    \hline
    \multicolumn{1}{|c}{Metrics}  & \multicolumn{3}{|c|}{Fenda} & \multicolumn{3}{c|}{Whale} \\ \cline{2-7}
   & CL.1 & CL.2& CL.3 & CL.1 & CL.2  & CL.3     \\ \hline
 Avg. \#Followers  &  627.6   & 749.5 & 951.4 & 508.7 & 1358.1 & 1687.1	\\ \hline
Avg. \#Listeners  &  16.6  & 27.0 & 25.9 & 43.5 & 68.1 & 62.6	\\ \hline
  Avg. Price (\$) &1.7  & 2.4 & 2.6 &0.2 & 1.3 & 0.7	\\ \hline
Avg. \#Questions &  106.5   & 68.8 & 71.4 & 41.6 & 17.7 & 112.4  \\\hline
 \end{tabular}
  \caption{User statistics of the identified clusters for
    Fenda and Whale.}
   \label{tbl:overallcompare}
  \vspace{-0.12in}
 \end{small}
\end{table}

 \begin{figure*}[t]
   \centering
   \centering
 \subfigure[Engagement.]
 {\label{fig:super:a}
     \includegraphics[width=0.32\textwidth]{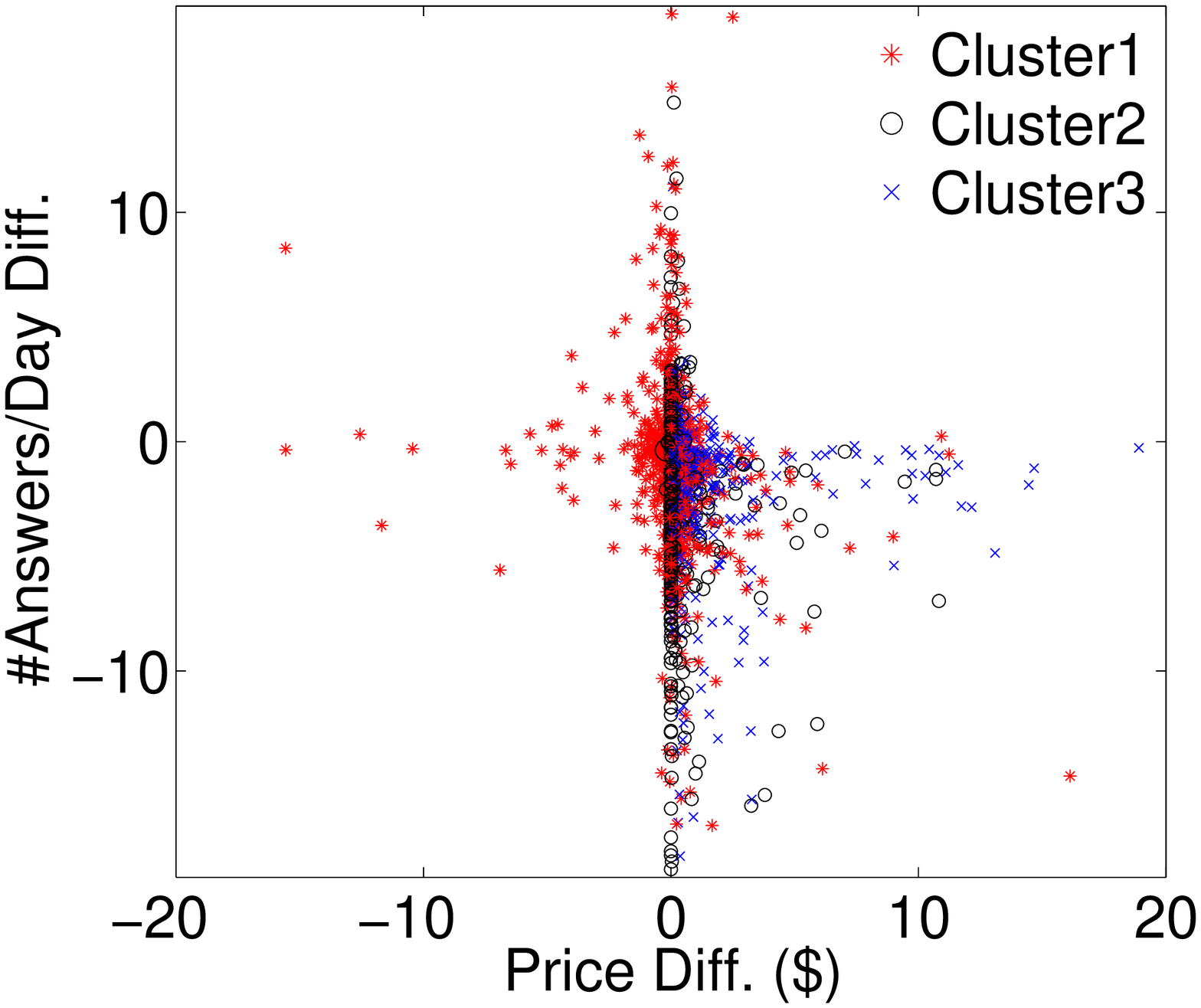}}
\hfill
  \subfigure[Income.]
  {\label{fig:super:b}
	\includegraphics[width=0.32\textwidth]{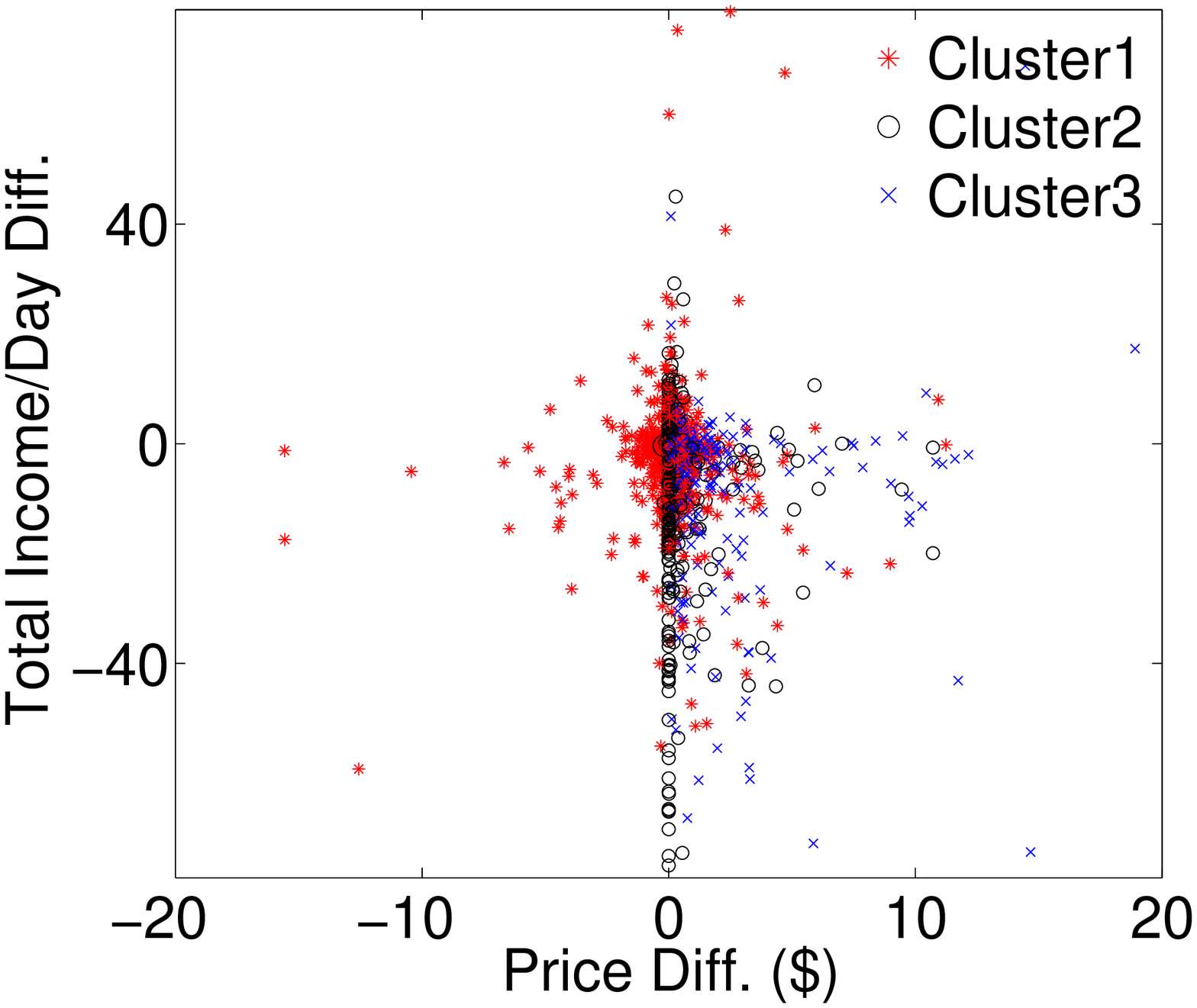}}
\hfill
  \subfigure[Listeners.]
  {\label{fig:super:c}
    \includegraphics[width=0.32\textwidth]{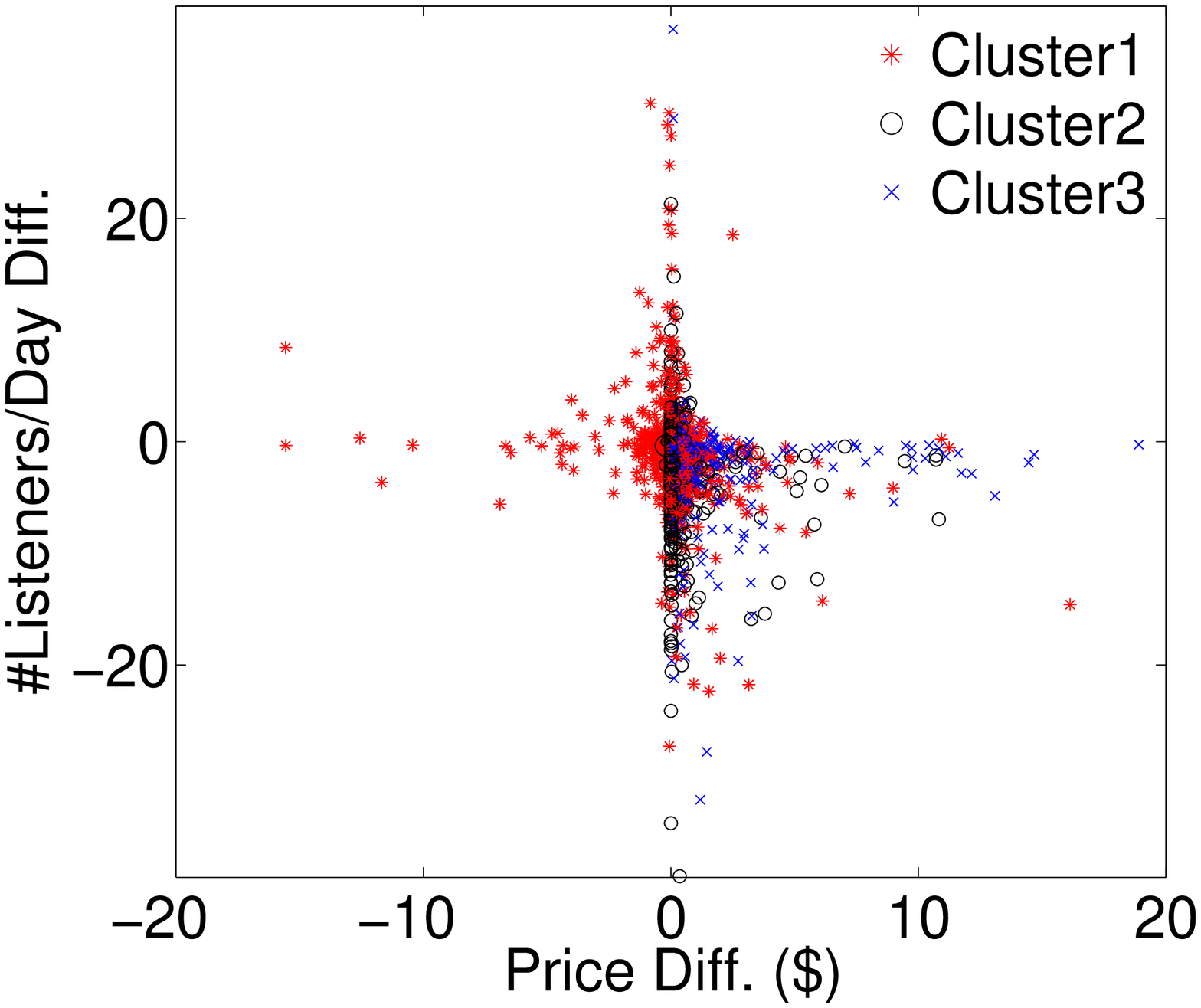}} 
    \vspace{-0.02in}
    \caption{The impact of pricing strategy to user engagement, income, and listeners over time in Fenda. We
      divide a user's lifespan in our dataset into two even parts, and
      compute the difference between the later half and the first
      half. A positive value indicates an upward trend (better viewed in color).}
        \vspace{-0.15in}
      \label{fig:compareengagement}
\end{figure*}



\subsection{Impact to User Engagement }
Next, we analyze how price adjustments affect a user's
engagement-level over time. Price is a key parameter within users'
control, and adjusting price is a way to test their answers' value in
the market. Intuitively, this price can affect a user's incoming
questions, earnings and social interactions, which are key incentive
factors to keep users engaged.  


\para{Fenda.} Figure~\ref{fig:super:a} shows the interplay between price
change and engagement level over time for 3 identified clusters on Fenda. We
quantify engagement-level using number of answers per day. To measure
changes over time, we divide a user's lifespan (time between her first
and last answer in our dataset) into two even parts. Then we compute
the differences for average price and engagement-level between the
later half and first half. In a similar way, we also measure the
changes in income (Figure~\ref{fig:super:b}) and listeners
(Figure~\ref{fig:super:c}), which represent the strength of monetary
and social incentives 

We observe different patterns: for cluster  2 and 3, more users are
located in the lower right corner than upper right, indicating a
decrease of engagement, income and number of listeners. A possible
explanation is that there is a mismatch between the answer's price and
its value, but users did not make the right adjustments. In contrast,
we find a significant number of users in cluster 1 located in the
upper left corner. By lowering their price, these users get to answer
more questions, and receive more money and listeners over time.
We validate the statistical significance of the results by calculating
the Pearson correlation~\cite{Sheskin:2007:HPN:1529939} between the
price change ($x$) and behavior metrics ($y$) for all three clusters
in Figure~\ref{fig:compareengagement}. We find 8/9 of the correlations
are significant ($p<0.05$) except for cluster1's income/day metric. 

Our result suggests that users need to set their price carefully to
match their market value. This requires proactive price adjustments
and lowering their price when necessary. Right now, highly
popular users on Fenda ({\em e.g.}, cluster 3) are less motivated or
unwilling to lower their price, which in turn hurts their income and
engagement level over time.

\para{Whale.} We performed the same analysis on Whale and found none
of the correlations were statistically significant (possibly due
to the small sample size). 

\section{Discussion}
Our analysis shows both positive impacts of monetary incentives and
some concerning issues in the long run. Below, we discuss the key
implications to future CQA design. 


First, {\em Answering On-demand.} Fenda and Whale adopt a
supplier-driven model where experts set a price for their answer. This
model is suitable for targeted questions (users know who to ask), but
can have a longer delay compared to crowdsourcing (where anyone can be
a potential answerer). Fenda and Whale achieve faster responses than most CQA services, but are still not as fast as
the crowdsourcing based Yahoo Answers. 
Recently, Fenda added a new
crowdsourcing channel for ``medical'' and ``legal'' questions. This
channel is customer-driven: users post their questions with a cash
reward, and any experts can give their answers to compete for the
reward. We did a quick crawling on the crowdsourcing channel and
obtained 1344 questions. We find their average response time is 4.38
hours, which is even faster than the 8.25 hours of Yahoo Answers
(Table~\ref{tbl:respondtime}). A promising future direction
is to explore a hybrid design of customer-driven and
supplier-driven model to further improve CQA efficiency. 




Second, {\em Rewarding Good Questions.} Fenda and Whale are the first
systems that reward users financially for asking good questions. This
leads to a mixed effect. On the positive side, users are motivated to
ask good questions that attract broad interests. 40\% of the questions
on Fenda received enough listeners to cover the asker's cost. Whale,
however, is less successful in profiting the askers due to the ``free
coin'' design. On the negative side, this model motivates
a small number of users to game the system for profits. We find
``bounty hunters'' who aggressively ask questions to low-priced
experts, and ``collusive'' users who work together to manipulate their
perceived popularity. Note that this is different from the traditional
cheating behavior in crowdsourcing platforms like MTurk, where
cheaters often produce low-quality work~\cite{Gadiraju:2015}. In Fenda
and Whale, manipulators mainly introduce unfairness, but they still
need to come up with good questions to attract listeners. 


Finally,  {\em Unfairness in Supplier-driven Markets.} In a
supplier-driven marketplace, a well-known expert has the key advantage
to receive questions. As a result, the financial income among
answerers is highly uneven: top 5\% answerers get about 90\% of the
total profits in Fenda. To attract questions, we find that less
popular users need to carefully adjust their price (including dropping
the price), while more popular users tend to increase their price. To
help users to bootstrap popularity, Fenda recently introduced a system
update, which allows users to set their answers ``free-for-listening''
for 30 minutes after posting. 

\section{Future Work and Limitations}
Fenda and Whale are among the recent wave of CQA systems that explore
a new design space for payment-based knowledge sharing
communities. They not only provide valuable
lessons for other services, but also raise new questions. 

\para{Communication Mechanisms for CQA.} Fenda and Whale let users record their answers in
  audio/video to avoid the inconvenience of typing text on the
  phone. Audio/video is likely to provide a more intimate
  experience for users, which, however, also makes it difficult to give longer answers. Future research may examine the proper communication channels (text, audio, video) for different
  Q\&A contexts. Noticeably, real-time streaming can be a promising
  option for CQA, given the huge success of Periscope and Facebook
  Live~\cite{Wang:2016imc}.


\para{Preventing Abuse and Manipulation.} Abusive activities are
likely to be a common problem for payment-based CQA given that
money is the incentive. Our work
provides a first look (empirically) at the bounty hunters and collusive
  users in Fenda and Whale. Future research is needed to develop more systematic
  approaches to detect abusive players and design new incentive models to
  prevent/limit abuse.

\para{Study Limitations.}
Our study has a few limitations. First, our study only focuses on
two services: Fenda and Whale. A broader comparison with other payment-based CQA
services can help to further generalize our results. Second, our dataset is not perfect. The crawler produces a
dataset with a complete list of experts but an incomplete list of normal
users. We argue that most of the missing users are likely lurkers (or inactive users) who are less influential in the community.  
We also used Fenda's official numbers to justify parts of our results. 
Finally, Fenda and Whale
are still exploring its way to building a sustainable CQA marketplace. It made a
few changes before our paper submission as discussed above. We
plan to continue to monitor these systems for future work. 

\section{Conclusion}
In this paper, we discuss lessons learned from the first
supplier-driven payment-based CQA systems. By analyzing a large
empirical dataset, we reveal the benefits of applying monetary
incentives to CQA systems (fast response, high-quality questions) as
well as potential concerns (bounty hunters and over time
engagement). As more payment-based CQA systems arise (Campfire.fm,
Quora Knowledge Prize, Zhihu Live), our research results can help system designers to make more informed design choices.

\balance
\begin{small}
\bibliographystyle{SIGCHI-Reference-Format}
\bibliography{mobility,biblio,astro}
\end{small}

\end{document}